**EyeCLIP: A visual–language foundation model for multi-modal ophthalmic image analysis**

Danli Shi, MD, PhD[1,2,#], Weiyi Zhang, MS[1,#], Jiancheng Yang, PhD[3], Siyu Huang, PhD[4], Xiaolan Chen, MD[1], Mayinuer Yusufu, MBBS[5], Kai Jin, MD, PhD[6], Shan Lin, MD[7], Shunming Liu, MS[8], Qing Zhang, MD, PhD[9], Mingguang He, MD[1,2,10]

**Author Affiliations:**

1. School of Optometry, The Hong Kong Polytechnic University, Kowloon, Hong Kong.

2. Research Centre for SHARP Vision (RCSV), The Hong Kong Polytechnic University, Kowloon, Hong Kong.

3. Swiss Federal Institute of Technology Lausanne (EPFL), Lausanne, Switzerland.

4. School of Computing, Clemson University, USA.

5. Centre for Eye Research Australia, Royal Victorian Eye and Ear Hospital, East Melbourne, Australia.

6. Department of Ophthalmology, The Second Affiliated Hospital, School of Medicine, Zhejiang University. China.

7. Bright Eye Hospital, Wuhan, China.

8. Department of Ophthalmology, Guangdong Academy of Medical Sciences, Guangdong Provincial People's Hospital, Guangzhou, China.

9. Beijing Tongren Eye Center, Beijing Tongren Hospital, Capital Medical University, Beijing, China

10. Centre for Eye and Vision Research (CEVR), 17W Hong Kong Science Park, Hong Kong.

[#] Contributed equally

**Correspondence**

**Prof. Mingguang He,** MD, PhD., Director of Research Centre for SHARP Vision; Chair Professor of Experimental Ophthalmology, School of Optometry, Kowloon, Hong Kong SAR. Email: mingguang.he@polyu.edu.hk

**Dr. Danli Shi,** MD, PhD., School of Optometry, The Hong Kong Polytechnic University, Kowloon, Hong Kong SAR, China. Email: danli.shi@polyu.edu.hk

**Abstract**

Early detection of eye diseases like glaucoma, macular degeneration, and diabetic retinopathy is crucial for preventing vision loss. While artificial intelligence (AI) foundation models hold significant promise for addressing these challenges, existing ophthalmic foundation models primarily focus on a single modality, whereas diagnosing eye diseases requires multiple modalities. A critical yet often overlooked aspect is harnessing the multi-view information across various modalities for the same patient. Additionally, due to the long-tail nature of ophthalmic diseases, standard fully supervised or unsupervised learning approaches often struggle. Therefore, it is essential to integrate clinical text to capture a broader spectrum of diseases. We propose EyeCLIP, a visual-language foundation model developed using over 2.77 million multi-modal ophthalmology images with partial text data. To fully leverage the large multi-modal unlabeled and labeled data, we introduced a pretraining strategy that combines self-supervised reconstructions, multi-modal image contrastive learning, and image-text contrastive learning to learn a shared representation of multiple modalities. Through evaluation using 14 benchmark datasets, EyeCLIP can be transferred to a wide range of downstream tasks involving ocular and systemic diseases, achieving state-of-the-art performance in disease classification, visual question answering, and cross-modal retrieval. EyeCLIP represents a significant advancement over previous methods, especially showcasing few-shot, even zero-shot capabilities in real-world long-tail scenarios.

**Introduction**

Ophthalmic diseases such as glaucoma, macular degeneration, and diabetic retinopathy pose a significant threat to global vision health, often leading to vision impairment or even blindness.[1] However, access to timely diagnosis and treatment remains a critical challenge due to insufficient medical resources, especially in underserved regions and developing countries.[2,3] This inequitable distribution of resources makes early detection and intervention for eye diseases particularly challenging, further exacerbating the burden of these diseases.

Artificial intelligence (AI) can significantly reduce specialists' workload by automating the analysis of ophthalmic images and assisting doctors in diagnosis.[4-7] Recently, the world has seen a shift from performing single tasks to building foundation models.[8-13] After pretraining on a large quantity of labeled or unlabeled data, the model can be easily adaptable to downstream tasks in a data-saving manner, reducing the cost and time of data preparation and improving the models' generalization capability. RETfound was the first proposed foundation model in ophthalmology using self-supervised reconstruction learning,[9] but it was trained on separate image modalities (color fundus photography [CFP] and optical coherence tomography [OCT]). To address this, EyeFound was proposed, which learns a shared representation of multi-modal ophthalmic imaging.[14] However, existing foundation models lack the modality-modality consistency and image-language alignment, which we believe is crucial in real-world scenarios.

In clinical practice, multiple examinations are optimal for examining different eye pathologies, such as CFP, OCT, fundus fluorescein angiography (FFA), and fundus autofluorescence (FAF).[15] Each examination provides unique and complementary information about the structure and function of the eye. Previous studies have demonstrated the complementary capabilities of different modalities in enhancing AI models for disease classification and segmentation.[16-20] Therefore, effectively utilizing multi-modal data is crucial for obtaining multi-view information, and ensuring consistency across modalities can serve as an important cue for self-supervised learning. Additionally, ophthalmic reports and diagnoses from experts' interpretation offer rich textual context, which should be assistive for learning tong-tailed representation with hierarchical concepts commonly encountered in the medical domain.[10,21] By integrating the clinical text, AI models can better simulate the cognitive processes of human experts, enabling them to handle complex, real-world clinical problems in an ever-changing environment.

In this work, we propose EyeCLIP, an ophthalmic visual-language foundational model designed to harness real-world multi-source, multi-modal data. EyeCLIP was pre-trained on a dataset comprising 2,777,593 multi-modal ophthalmic images and 11,180 reports from 128,554 patients using self-supervised learning and multi-modality alignment. Specifically, the training combined self-supervised reconstruction, multi-modal image contrastive, and image-text contrastive learning. Subsequently, we validated EyeCLIP on 14 multi-country datasets to assess its performance in zero-shot, few-shot, and supervised settings across different tasks, including multi-modal ocular disease diagnosis and systemic disease prediction, visual question answering (VQA), and cross-modal retrieval. EyeCLIP can effectively learn a shared representation of multiple examinations, enabling zero-shot disease diagnosis and improved language understanding by fully utilizing a large amount of unlabeled, multi-examination, and labeled data in the real world. We believe our approach not only represents a significant advancement in ophthalmic foundation models but also offers insights for training foundational models with incomplete multi-modal medical data accumulated in clinical practice across other medical domains.

**Results**

**EyeCLIP development using multi-center multi-modal datasets**

The EyeCLIP system was trained using 2,777,593 multi-modal images and 11,180 reports from 128,554 patients across diverse regions and hospitals in China to learn ophthalmic vision-language features comprehensively. The data details can be found in Figure 1 and the Methods section. Following training, EyeCLIP can be directly applied to applications involving classification and cross-modal retrieval without further training. Also, it can be finetuned in a data-saving manner for downstream applications such as ocular disease diagnosis, systemic disease prediction, and interactive VQA. Figure 1 shows the study design. The characteristics of the 14 downstream datasets can be found in Extended Table 1. Figure 2a presents EyeCLIP's overall superior performance across different downstream tasks compared with the general-domain CLIP model,[22] medical domain BioMedCLIP,[23] PubMedCLIP,[24] and ophthalmology domain RETFound[9].

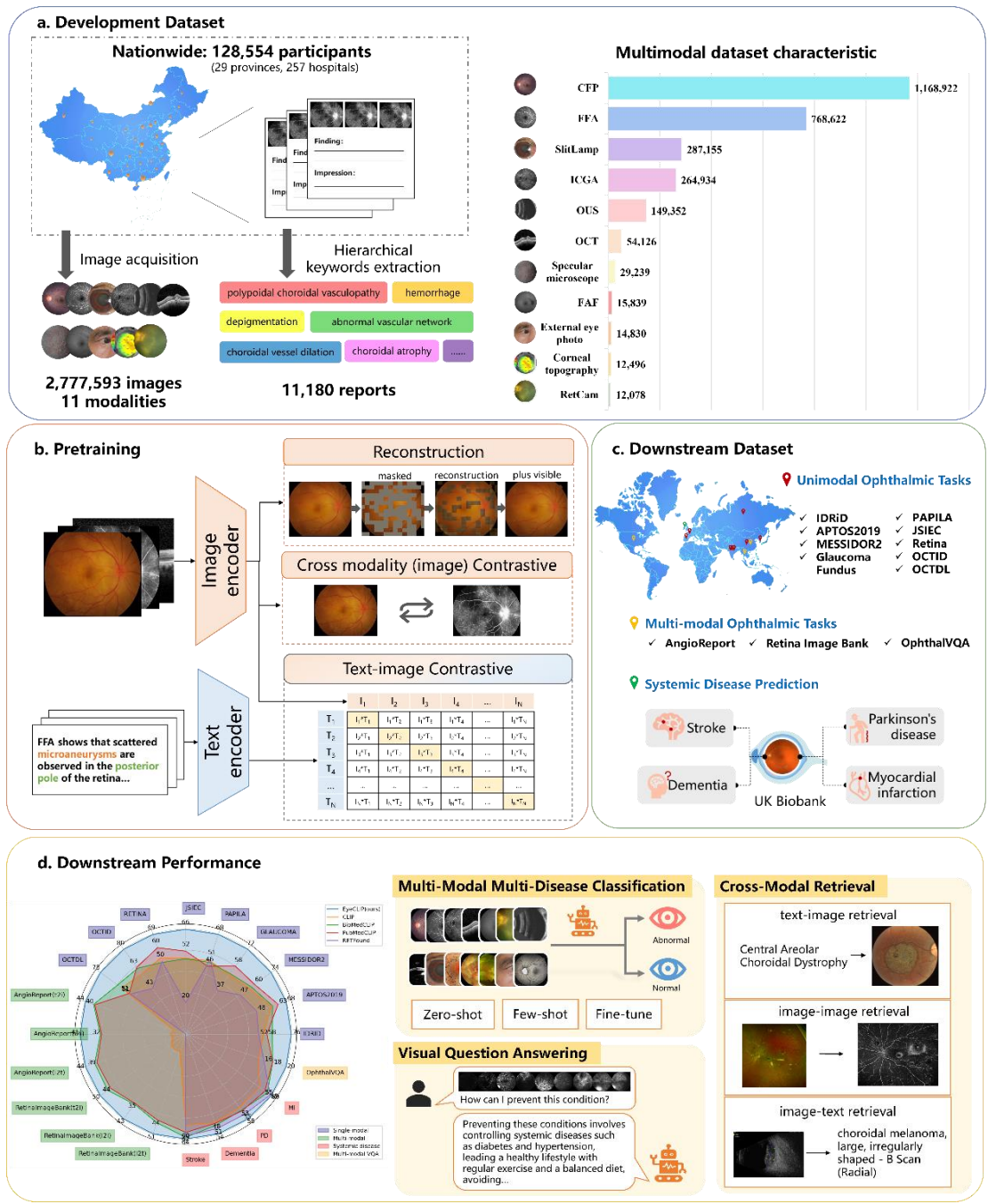

**Figure 1**. Study diagram. a. Using an extensive multi-modal database across nine provinces in China, we matched the multi-examination images from the same patient, and cleaned the medical reports using a keyword mapping dictionary containing medical terminology to generate hierarchical keyword text labels. b. EyeCLIP was pretrained using self-supervised reconstruction, multi-examination contrastive learning, and hierarchical text-image contrastive learning to fully leverage real-world multi-examination clinical data. c. Downstream multi-country datasets for EyeCLIP validation, including zero-shot, few-shot, and supervised finetuning scenarios. d. Radar plot outlines the performance of EyeCLIP and baseline models across various downstream tasks. EyeCLIP significantly outperforms the baseline models across diverse tasks, including zero-shot classification, multi-modal retrieval, visual

question answering (VQA), and supervised systemic disease prediction.

**EyeCLIP excels in zero-shot, partial and full-data training ocular disease classification**

Zero-shot transfer capability enables a single pretrained foundation model to be applied directly to downstream tasks. EyeCLIP could be a strong baseline for conventional supervised learning, especially when training labels are scarce. We evaluated EyeCLIP's zero-shot classification performance without task-specific training on nine public ophthalmic datasets. Using CFP as the input modality, EyeCLIP significantly outperformed other models in diagnosing ophthalmic diseases (all P<0.001), with AUCs ranging from 0.681 to 0.757 for DR, 0.721 and 0.684 for glaucoma, as well as 0.660 and 0.688 for multi-disease diagnosis. For OCT, EyeCLIP achieved the highest AUROC scores of 0.800 for OCTID[25] and 0.776 for OCTDL[26], higher than the other models (all P<0.001). Quantitative results are presented in Figure 2b and Extended Table 2.

Next, we evaluated the few-shot performance of EyeCLIP on those nine ocular disease datasets, using limited training samples of 1, 2, 4, 6, and 16, respectively. The results indicated that EyeCLIP could generalize with limited data, demonstrating the ability to diagnose various ophthalmic diseases data-efficiently, outperforming other models (all P<0.01). Quantitative results of AUROC and AUPR are provided in Figure 3 and Extended Table 3.

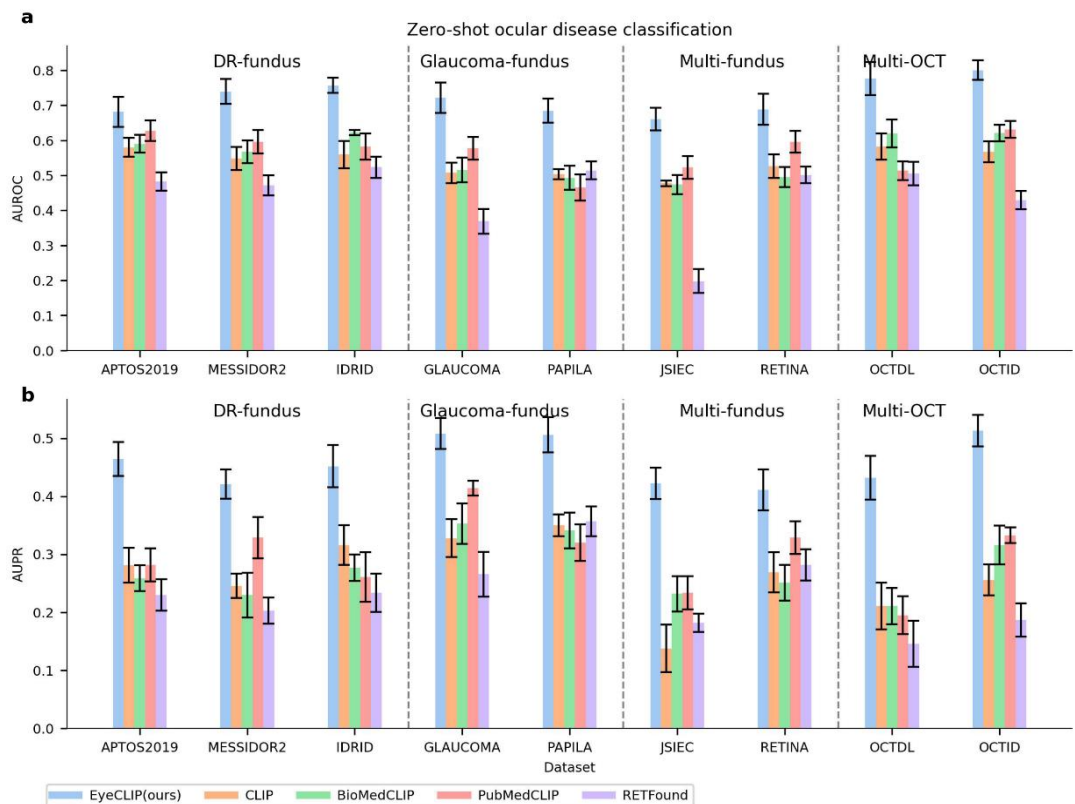

**Figure 2**. Zero-shot performance on downstream ocular diseases datasets. a. AUROC. b. AUPR. Error bars represent 95% confidence intervals, and the centers correspond to computed values of each metric. EyeCLIP achieved significantly better zero-shot performance than other models for both AUROC and AUPR. AUROC = area under the receiver operator characteristic curve, AUPR = area under the precision-recall curve.

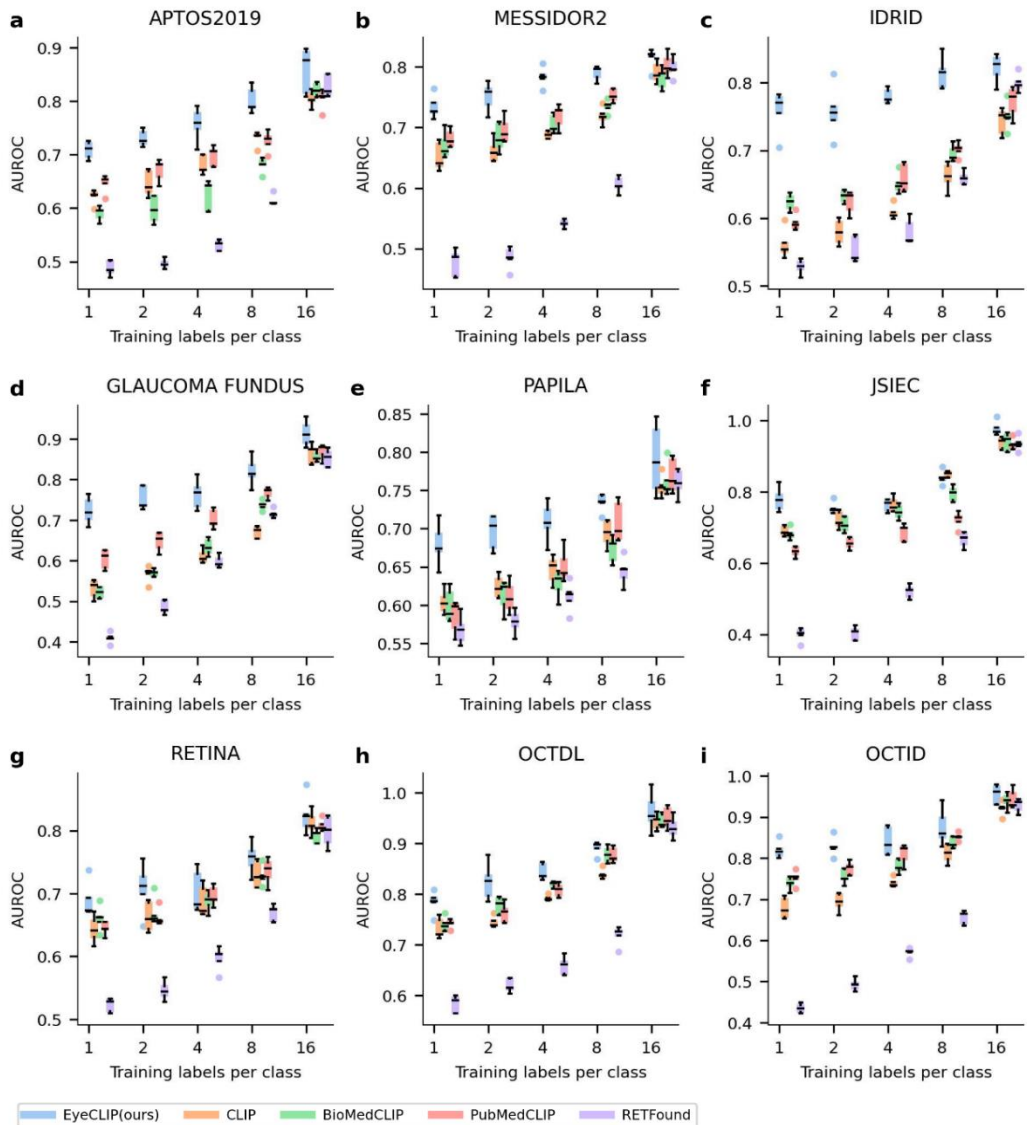

**Figure 3**. Few-shot classification experiments. We investigated the label efficiency of different pretrained models in a few-shot setting, varying the number of training labels per class (nc = 1, 2, 4, 8, 16). For each nc, we sampled five different sets of training examples and trained a weakly supervised model. Boxes indicate quartile values, and whiskers extend to data points within 1.5× the interquartile range. EyeCLIP achieves significantly better performance (in terms of the mean AUROC of five runs) than other encoders for different sizes of training sets and across all datasets. AUROC = area under the receiver operator characteristic curve. AUPR results can be found in Extended Figure 1.

Specifically, rare diseases are known for lacking sufficient data due to low incidence rate, and they are common challenges medical AI, and should be most beneficial with data-efficient training. Therefore, we further evaluated its performance for few-shot classification using a subset of Retina Image Bank selected by ophthalmologists, with the number of images for each class larger than 16. The subset included 17 rare diseases: acute posterior multifocal placoid pigment epitheliopathy, birdshot retinochoroidopathy, central areolar choroidal dystrophy, choroidal melanoma, choroidal osteoma, cone dystrophy, congenital hypertrophy of the retinal pigment epithelium, familial exudative vitreoretinopathy, macular telangiectasia, optic disc pit, optic nerve hypoplasia, pseudoxanthoma elasticum, retinitis pigmentosa, retinoblastoma, retinopathy of prematurity, serpiginous choroiditis, Stargardt disease. EyeCLIP beat other models in classifying rare diseases in all settings. The results are presented in Figure 4c and Extended Table 4.

Lastly, we tested EyeCLIP using the full-data supervised training paradigm on 11 public datasets containing unimodal and multi-modal images, with a train, validation and test split ratio of 55:15:30%. Detailed results are provided in Figure 4a and Extended Table 5.

For single-modality task, EyeCLIP outperformed competing models except for three datasets when it was on par with the 2$^{nd}$ best model RETFound. In DR classification, EyeCLIP significantly surpassed RETFound in IDRiD dataset [with AUROC 0.835 vs 0.826, P=0.013], which is a small dataset, but on par with RETFound on much larger datasets APTOS2019 and MESSIDOR2 (P>0.05), suggesting EyeCLIP surpasses RETFound in a matter of data efficiency, requiring less data than RETFound. For glaucoma and multi-disease classification, EyeCLIP consistently outperformed other models. For OCT images, EyeCLIP was on par with RETFound on OCTID dataset (P>0.05), but significantly better on OCTDL dataset (AUROC 0.993 vs 0.982, P<0.001), which is a more imbalanced dataset with long-tailed classes. Even though RETFound specifically trained separate weights that are optimal for CFP and OCT, EyeCLIP is generally better and no worse than it, even with one general encoder.

For multi-modality tasks, EyeCLIP outperformed all comparison models. On the AngioReport (APTOS2023[27]) dataset with two modalities, EyeCLIP outperformed the next best model BioMedCLIP with an AUROC of 0.721 versus 0.705, P<0.001. Moreover, EyeCLIP performed the best on the challenging Retina Image Bank[28] dataset with 14 modalities and 84 conditions including rare diseases, with AUROC of 0.561 versus the 2$^{nd}$ best 0.545, P<0.001.

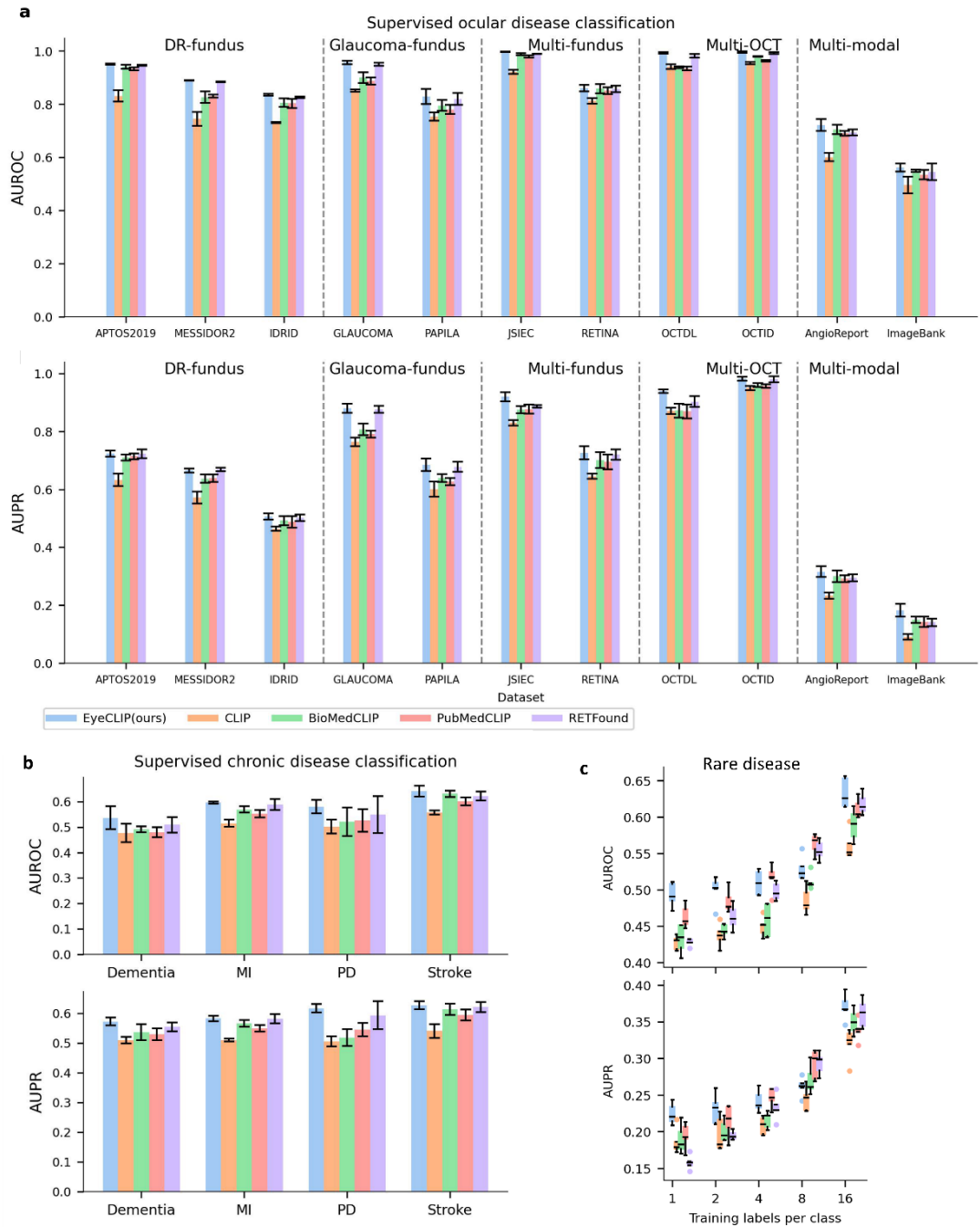

**Figure 4**. a. Supervised full-data finetuning on ocular disease tasks. EyeCLIP is on par with the 2nd best model RETFound on APTOS2019, MESSIDOR2, OCTID (P>0.05), and surpasses all models on the other eight datasets. b. Supervised full-data finetuning on systemic disease prediction. EyeCLIP surpasses all other models. (P<0.05). c. Few-shot finetuning on rare disease classification. EyeCLIP surpasses all other models. (P<0.05). Boxes indicate quartile values, and whiskers extend to data points within 1.5× the interquartile range. Detailed statistics can be found in Extended Data Tables 4-5. AUROC = area under the receiver operator characteristic curve, AUPR = area under the precision-recall curve.

**EyeCLIP enhances systemic disease prediction**

Systemic diseases such as stroke and myocardial infarction (MI) pose significant threats to older adults, often leading to sudden death. The eyes, rich in blood vessels that can be directly visualized, have been referred to as "the window to the health of the body."[29] Therefore, predicting the incidence of systemic diseases is a crucial technique for early screening and prevention. However, compared to the general population, the incidence of these events is relatively low, resulting in limited positive training data. Consequently, data-efficient training methods are highly valued in this context. We evaluated EyeCLIP's performance in predicting systemic diseases based on ophthalmic images using the UK Biobank[30]. Our experiment included predictions for stroke, dementia, Parkinson's disease (PD), and MI. We first assessed the few-shot performance of EyeCLIP using limited training samples of 1, 2, 4, 6, and 16, respectively. EyeCLIP consistently outperformed other models, demonstrating superior data efficiency in predicting systemic diseases. For full-data supervised training, EyeCLIP ranked first, achieving AUROC scores of 0.641, 0.536, 0.580, and 0.596, and AUPR scores of 0.627, 0.572, 0.616, and 0.582, respectively (all $P<0.05$). Detailed results are provided in Figure 4b and Extended Tables 6-7.

**EyeCLIP achieves zero-shot cross-modal retrieval**

By learning an aligned latent space for multi-modal embeddings, EyeCLIP enabled zero-shot cross-modal retrieval. This included retrieving text entries based on image queries (image-to-text, i2t), retrieving images based on text queries (text-to-image, t2i), and retrieving images based on image queries (image-to-image, i2i). This function is useful for biomedical applications such as identifying cases for research cohorts, assisting with rare disease presentations, and creating educational resources. We evaluated EyeCLIP on two external multi-modal image-caption datasets, AngioReport and Retina Image Bank, which cover a diverse range of ophthalmology concepts. To specifically investigate the performance on rare diseases, we manually selected a subset from Retina Image Bank containing only rare diseases. Following previous studies,[31,32] we used Recall@K as the metric for cross-modal retrieval.

On AngioReport, EyeCLIP achieved mean recall of 44.1%, 40.7%, and 44.3% for text-to-image, image-to-image, and image-to-text retrieval, respectively, outperforming BioMedCLIP's 40.5%, 32.9%, and 40.1% ($P < 0.01$ for all tasks). On Retina Image Bank, EyeCLIP achieved mean recall of 50.2%, 43.3%, and 50.9%, outperforming BioMedCLIP's 45.8%, 35.8%, and 45.3% ($P < 0.01$ for all tasks). Extended Table 8 presents the details of model performance. Examples of the retrieved results are presented in Figure 5; EyeCLIP effectively retrieved similar contents using text or images as queries. It could retrieve relevant images based on text descriptions, pair images with the same pathological condition or from the same patient, and the most correlated description with the image inputs.

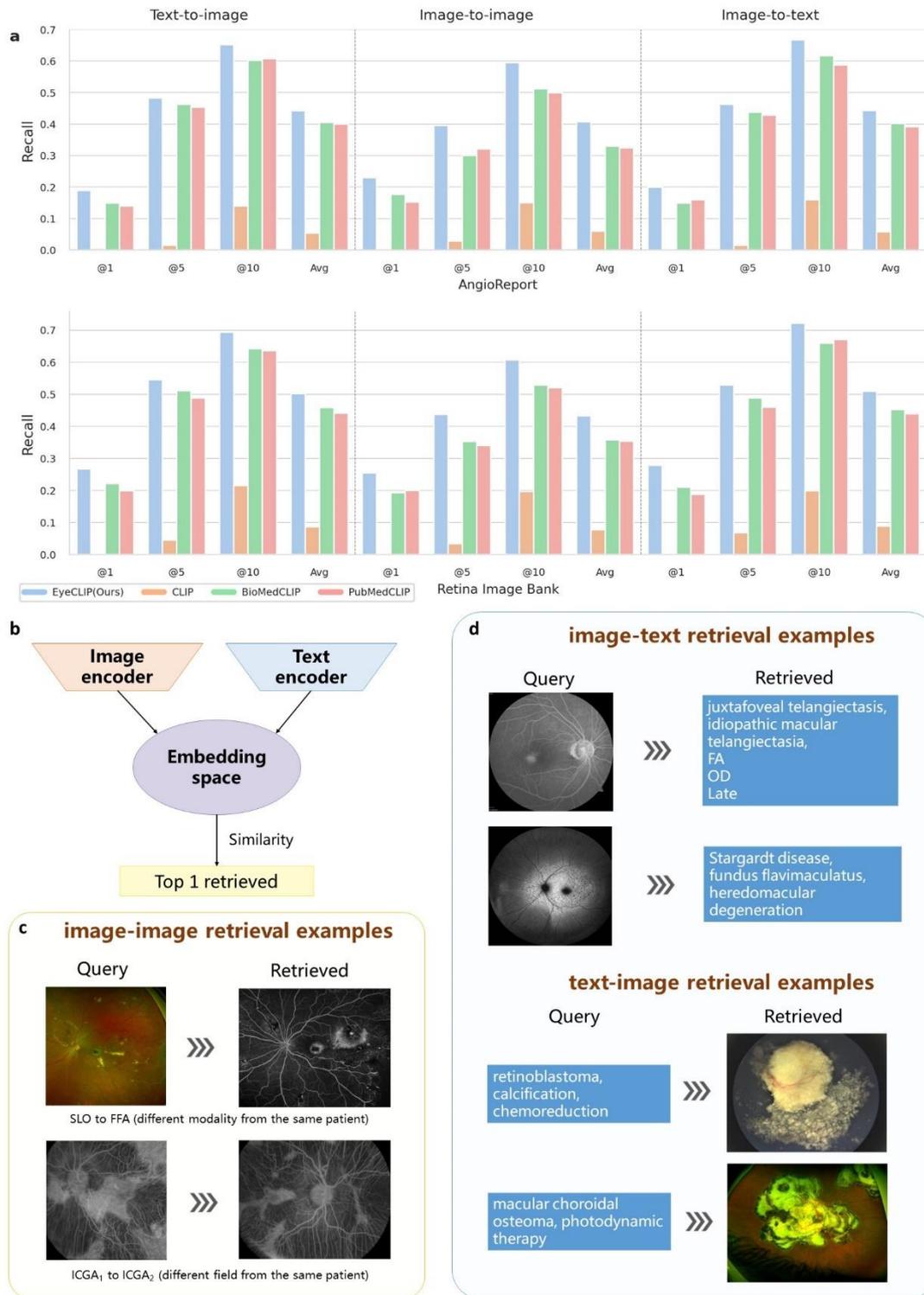

**Figure 5**. Zero-shot multi-modal retrieval performance. a. Model comparison on two datasets with image-text pairs, AngioReport and Retina Image Bank. Similarity in the embedding space was computed between the query image and all text samples in the database. The top-K most similar texts were retrieved. We report Recall@K for K ∈ {1, 5, 10} and the mean recall, which averages over K. We compared different models in text-to-image (1st column), image-to-image (2nd column) and image-to-text (3rd column). EyeCLIP outperforms other baselines on all retrieval tasks. Error bars indicate

95% confidence intervals. b, Schematic illustrates zero-shot cross-modal retrieval. c & d. Examples of images in the top one retrieved result from the Retina Image Bank. More examoles can be found in Extended Figure 2.

**EyeCLIP demonstrates zero-shot generalization on VQA**

Vision-language foundation models hold the potential for generalization in ophthalmic VQA. We combined the image encoder from each model and a text encoder with a large language model (LLM): Llama2-7b to perform VQA. Since there lacks a well-established public ophthalmic VQA dataset to perform few-shot and full-data finetuning experiments, we performed finetuning using the Retina Image Bank multi-disease data with generated Q of "Diagnosis:" to align the image, question, and LLM features. Subsequently, we performed zero-shot VQA on the external OphthalVQA[33] dataset without further training. OphthalVQA is an open-set VQA dataset containing 60 images in 6 modalities with 60 ophthalmic conditions and 600 QA pairs. As shown in Extended Table 9, EyeCLIP demonstrated superior alignment with LLM and could achieve VQA despite the image and language modules were not specifically aligned on open-set VQA data, EyeCLIP ranked first across all metrics, including exact matching score, F1 score, Precision, Recall, and Bilingual Evaluation Understudy (BLEU)[34] (P<0.001).

**Discussion**

In this study, we developed EyeCLIP, a visual-language foundation model for multi-modal ophthalmic image analysis, utilizing a large dataset of 2,777,593 ophthalmic images spanning 11 modalities, along with corresponding hierarchical language data. Our novel training strategy fully leverages real-world data nature, characterized by multi-examination and large amounts of unlabeled and labeled data. This approach achieved a shared representation across multiple examinations and modalities. EyeCLIP significantly enhances the analysis of ophthalmic and systemic diseases, demonstrating state-of-the-art efficiency and generalizability in zero-shot, few-shot, and full-data finetuning downstream tasks.

One primary advantage of EyeCLIP lies in its alignment of multi-examinations which is demonstrated in the image-image retrieval task and multimodal image classification tasks. In contrast, conventional foundation models often focus on specific types of examination, which limits their effectiveness for real-world applications. Given the complexity of real-world clinical settings, where patients present with various conditions and undergo multiple tests, a model capable of accurately identifying diverse eye conditions with different image modalities is highly desirable. The development of EyeCLIP involves 11 modalities from diverse populations makes unique and powerful, demonstrated performance in identifying vision-threatening diseases, particularly in multi-modal, multi-disease diagnostics with label imbalance. Notably, the challenging Retina Image Bank underscores its potential for managing rare eye conditions with diverse examination.

Another major strength of EyeCLIP is its integration of visual-language pretraining. While previous foundation models primarily focused on extracting meaningful patterns from rich image data, EyeCLIP utilized textual descriptions created by medical professionals to distill hierarchical context information. By employing text-image contrastive learning, EyeCLIP maximized the use of all available labeled ophthalmic data, learning semantically rich features of disease manifestation. This alignment offers zero-shot capabilities, significantly reducing the need for extensive annotation of training data. This feature is especially beneficial in resource-limited settings and remote areas, where access to specialized medical care is constrained. And for rare diseases. Moreover, the zero-shot VQA capability presents a unique opportunity to automate interpretative tasks in clinical settings with minimal model adjustments. EyeCLIP's ability to operate with minimal training data and adapt to new tasks makes it a valuable tool for expanding the reach of quality ophthalmic care widely.

Ophthalmic images are increasingly used to indicate systemic diseases due to their accessibility.[35,36] This is an area where the foundation model could be well appreciated due to the scarcity of events data compared with a healthy population. Notably, EyeCLIP significantly improved systemic disease prediction, surpassing previous medical domain foundation models like BioMedCLIP and ophthalmology domain RETFound in events including stroke, dementia, PD, and MI. This improvement is likely attributed to the shared representation of different examination data. For example, angiography provides better visualization of retinal blood vessels and lesions, and these features could be jointly learned by the model. After further optimization, EyeCLIP can be a powerful tool for early detection and monitoring of systemic diseases, enhancing patient care beyond ophthalmology.

This study offers valuable insights for other medical domains dealing with incomplete or unaligned data. In real-world clinical practice, it is common for datasets to contain multi-modal information, such as images and text, that are not fully aligned across every sample. In this work, we address this challenge by employing a strategy that combines self-supervised learning through masked-image reconstruction within single modalities and contrastive learning across aligned multi-modal data when available. This approach maximizes the utility of diverse clinical data accumulated in practice, offering a potential framework for developing medical foundation models in other fields where incomplete multi-modal data is prevalent.

Our study has several limitations. Firstly, EyeCLIP's performance relies on the quality and diversity of the training data. Additional training using more comprehensive clinical and demographic datasets with more text labels may improve its predictive performance and practicality in different populations. Secondly, integrating vision and language data poses challenges. The quality and consistency of language descriptions vary depending on the expertise and documentation practices of medical professionals. Developing standardized protocols for generating and annotating textual data in ophthalmology and implementing structured reporting templates to ensure uniformity is expected to mitigate this issue and enhance the model's learning from multi-modal data. Thirdly, deploying EyeCLIP in real-world clinical settings requires careful consideration of practical and ethical issues. The model's predictions need to be interpretable and transparent to gain the trust of healthcare providers and patients, ensuring successful implementation in clinical practice.

In conclusion, we developed EyeCLIP, a visual-language foundation model characterized by shared multi-modal representations capable of performing a wide range of downstream tasks. The novel training strategy aligns well with real-world data characteristics, potentially informing the development of foundation models in general medicine. EyeCLIP's outstanding performance and broad applicability to ocular and systemic diseases position it as a promising tool to enhance the accuracy, efficiency, and accessibility of AI in ophthalmic clinical practice and research.

**Methods**

**Ethics Statement**

This study was conducted in accordance with the Declaration of Helsinki and received approval from the Hong Kong Polytechnic University's institutional review board (HSEARS20240202004). The IRBs waived informed consent due to the retrospective analysis of anonymized ophthalmic images and public datasets.

**Data Curation and Preprocessing for Pretraining**

We collected a vast amount of unlabeled ophthalmic images from 227 hospitals across China, totaling 2,777,593 images from 128,554 patients. These images covered a variety of ocular conditions and comprised 11 different image modalities, including CFP, FFA, indocyanine green angiography (ICGA), and OCT, among others. To ensure the quality of the data, we excluded low-quality images from CFP, FFA, and ICGA by extracting and analyzing the vascular structures. Specifically, images with detachable vascular ratios less than 0.04 for CFP and less than 0.01 for FFA and ICGA were removed. The language training data were sourced from 11,180 angiography reports of 11,180 participants. Since the reports contain custom templates and are generally long, we developed a custom dictionary and employed hierarchical keyword extraction algorithms to convert medical knowledge from the reports into a set of keywords covering various aspects such as ophthalmic diseases, anatomical structures, and diagnostic indicators.[7] This process provided crucial semantic information for subsequent image-text alignment and pre-training. Before model development, all data, including images and ophthalmic reports, were de-identified. Additional information about the pretraining dataset is summarized in Figure 1.

To facilitate multi-modality alignment, we matched ophthalmic images of different examinations to obtain image pairs from the same patient, enabling the model to learn features across different imaging examinations better. Since the reports often contain templates with long redundant information, we cleaned the medical reports using a keyword mapping dictionary[6] containing medical terminology to generate hierarchical keyword text labels.

**Data Curation and Preprocessing for downstream validation**

Extended Table 1 summarizes the details of datasets used for downstream validation. We included 14 datasets, covering ocular disease diagnosis (multi-class classification), systemic disease prediction, and multi-modal disease classification (multi-label classification), and VQA.

*Ophthalmic single-modality classification datasets*

We organized 9 publicly available single-modality ophthalmic disease classification datasets from different ethnicities and regions, including 7 CFP and 2 OCT datasets. The CFP datasets included IDRiD (India, 516 images),[37] APTOS2019 (India, 3,662 images), and MESSIDOR2 (France, 1,744 images) for DR diagnosis; PAPILA (Spain, 488 images)[38] and Glaucoma Fundus (South Korea, 1,544 images)[39] for glaucoma diagnosis; as well as JSIEC[40] and Retina for the classification of multiple ophthalmic diseases. The OCT datasets included OCTID (India, 572 images)[25] and OCTDL (Russia, 2064 images)[26], both containing multiple disease labels.

*Ophthalmic multi-modality classification datasets*

We also collected two multi-modality, multi-label datasets: the AngioReport[27] dataset and the Retina Image Bank[28]. The AngioReport dataset consists of approximately 50,000 angiographic images collected from routine clinics in Thailand, including FFA and ICGA modalities and covering 142 retinal diseases. We selected a test subset of 10,520 images to validate our model. The Retina Image Bank, sourced from the United States, is a large open-access repository of retinal images containing 14 modalities and 84 ophthalmic diseases. We obtained images and their corresponding findings from the website and created a custom dictionary to standardize different disease expressions using keyword matching and regular expressions. The standardized labels incorporate hierarchical structures, such as "DR, mild DR" for mild diabetic retinopathy. We excluded non-standard retinal examination images, including schematic cartoons, histology, and pathology images. To increase efficiency, we focused on images uploaded between 2019 and 2023 and removed instances with fewer than 50 occurrences. This process yielded a final dataset of 3,293 images.

*Ophthalmic VQA Dataset*

OphthalVQA[33] is a multi-modal ophthalmic image-QA dataset from China, comprising 60 images across six modalities: slit-lamp, scanning laser ophthalmoscope (SLO), CFP, OCT, FFA, and ocular ultrasound images. It contains 60 ophthalmic conditions with 600 human-curated VQA pairs. The VQA pairs covered common questions related to modality recognition, disease diagnosis, disease examination, and treatment.

*Systemic Chronic Disease Dataset*

UK Biobank[30] is a population-based prospective cohort from the United Kingdom, recruiting approximately 500,000 participants aged 40 to 69 between 2006 and 2010. Most participants have extensive phenotypic data. We utilized CFP images and data on systemic diseases. Using algorithm-defined outcomes (category 42), we predicted four major systemic diseases that significantly impact health: stroke, dementia, PD, and MI based on CFP images. To avoid potential bias from inconsistent individual visits, we included only retinal images of the right eye from a single visit per patient.

**Model Design and Training Details**

All experiments were conducted in Python 3.10. For visual-language pretraining, we employed CLIP[22] as our base framework, which is a pretrained model that leverages contrastive learning using image-text pairs. This model processes image and text inputs independently through an image encoder and a text encoder, generating distinctive vector representations for each modality. Subsequently, these vectors are projected into a unified multi-modal embedding space, facilitating direct comparisons between textual and visual elements.

We extended the traditional CLIP architecture by adding an image decoder to the CLIP image encoder, following Masked Autoencoders (MAE).[41] This addition enables the model to perform masked image reconstruction, pivotal for self-supervised feature representation learning. Specifically, besides the original image-text contrastive loss $\mathcal{L}_{img-text}$, we modified the loss function of CLIP by adding an image reconstruction loss $\mathcal{L}_{recon}$, and an image-image contrastive loss $\mathcal{L}_{img-img}$.

$\mathcal{L}_{img-text}$ is used to align the image and corresponding text descriptions, which is defined as:

$$\mathcal{L}_{img-text} = -\frac{1}{N}\sum_{i=1}^{N} \log \frac{\exp(\text{sim}(f(x_i), g(t_i))/\tau)}{\sum_{j=1}^{N} \exp(\text{sim}(f(x_i), g(t_j))/\tau)}$$

where $f(x)$ and $g(t)$ are the encoded image and text representations, $sim$ denotes the similarity measure, typically cosine similarity, and $\tau$ is a temperature parameter.

Similarly, $\mathcal{L}_{img-text}$ aligns the features between different modalities of images, which is defined as:

$$\mathcal{L}_{img-img} = -\frac{1}{N}\sum_{i=1}^{N} \log \frac{\exp(\text{sim}(f(x_i), f(x_j))/\tau)}{\sum_{k=1}^{N} \exp(\text{sim}(f(x_i), f(x_k))/\tau)}$$

$\mathcal{L}_{recon}$ is the loss for reconstructing masked images, which is defined as:

$$\mathcal{L}_{recon} = \frac{1}{N}\sum_{i=1}^{N} |\hat{x}_i - x_i|_2^2$$

Where $\hat{x}$ and $x$ are the reconstructed and original images, respectively.

The final loss function for training our model is the combination of the three losses:

$$\mathcal{L} = \lambda_{\text{img-text}}\mathcal{L}_{img-text} + \lambda_{\text{img-img}}\mathcal{L}_{img-img} + \lambda_{\text{recon}}\mathcal{L}_{recon}$$

Among them, $\lambda_{img-text}$ and $\lambda_{img-img}$ are set to 0.75, and $\lambda_{recon}$ is set to 1.

In EyeCLIP, all images share the same encoder, ensuring consistent feature extraction across different modalities. This innovative combination of CLIP and MAE distinguishes our approach from traditional CLIP models, enhancing its capability by utilizing the large amount of unlabeled data.

During the training phase of EyeCLIP, we cropped the images to field-of-view and resized them to 224 × 224. The EyeCLIP was trained with a base learning rate of 0.001 for the first 2,000 steps, with a 2-epoch warm-up, followed by cosine decay to zero throughout the training process. A batch size of 200 was utilized, and training was conducted on one NVIDIA Tesla V100 (32GB) GPU for approximately four weeks. At the end of the training, the model with the lowest loss on the validation set was selected for testing.

**Details of the Comparison Models**

PubMedCLIP

PubMedCLIP is a CLIP model specifically fine-tuned for the medical domain.[24] Trained on the Radiology Objects in COntext (ROCO) dataset,[42] it encompasses over 80,000 samples from various medical imaging modalities like ultrasound, X-rays, computed tomography, magnetic resonance imaging, and various body regions. The texts used for training were the relatively short captions associated with the images in the ROCO dataset. Experimental outcomes showcased that leveraging PubMedCLIP as a pre-trained visual encoder led to a potential performance boost of up to 3% for existing MedVQA models.

BioMedCLIP

BioMedCLIP is a multi-modal biomedical foundation model pre-trained using 15 million scientific image-text pairs extracted from 4.4 million articles in PubMed Central.[23] It incorporates a domain-specific language model (PubMedBERT),[43] utilizes larger vision transformers, and integrates other domain-specific optimizations. Compared to general-domain CLIP and previous biomedical vision-language models such as PubMedCLIP, BioMedCLIP demonstrates superior performance across various downstream tasks, including cross-modal retrieval, zero-shot image classification, and VQA.

RETFound

As the first foundation model in the field of ophthalmology, RETFound[9] is trained on a vast dataset comprising 1.6 million unlabeled retinal images through self-supervised reconstruction. It leveraged two ophthalmic modalities, CFP and OCT, to train separate weights for each modality. RETFound surpassed other comparative models, including those pretrained on ImageNet, in diagnosing sight-threatening eye conditions, and the prediction of systemic disorders.

**Classification Downstream Tasks**

Zero-shot Classification

For zero-shot transfer, we followed the method in the CLIP experiment. Each class was associated with a text prompt consisting of the modality and class name (for example, 'color fundus, diabetic retinopathy'). We computed the $\ell2$-normalized embedding using the text encoder and image encoder from EyeCLIP for the prompt and image. For each image, we computed the $\ell2$-normalized embedding and then computed cosine-similarity scores between the image and each text embedding, and the predicted class was consequently the class with the highest similarity score.

Full-data Fine-tune Classification

We used each image encoder to extract a low-dimensional feature embedding from each image and added a multilayer perceptron to map the image feature representation to logits, which were interpreted as class probabilities after softmax normalization. During finetuning, the encoder was frozen for the first five epochs and unfrozen afterward. A total of 50 epochs was trained for each model. For single-label classification tasks, we used a batch size of 16. The first ten epochs implemented a learning rate warm-up from 0 to $5 \times 10^{-4}$, followed by a cosine annealing schedule reducing the learning rate from $5 \times 10^{-4}$ to $1 \times 10^{-6}$ over the remaining 40 epochs. For multi-label classification tasks in AngioReport and Retina Image Bank, we used a batch size of 4, trained for 30 epochs, and set the learning rate to 0.01. After each epoch, we evaluated the model on the validation set, saving the model weights with the highest AUROC for internal and external assessments.

Few-shot Classification

We varied the number of labeled examples per class for finetuning EyeCLIP (known as 'shot') from n = 1, 2, 4, 8, 16, and tested the model on the test set similar to full-data finetune classification.

Cross-Modal Retrieval

For cross-modal retrieval, we used the same method as zero-shot classification above to retrieve the top-K images that were closest in the aligned latent space to a specific text query (text-to-image retrieval). Image-to-text and image-to-image retrieval were performed analogously. To evaluate retrieval, we used Recall@K, which measures the percentage of correct results included in these top-K retrieved samples. We chose K ∈ (1, 5, 10) and reported mean recall by averaging the scores over the three Recall@K values.

Visual Question Answering

We used the image encoder from EyeCLIP to extract image features, which were then concatenated with text features (questions). The combined feature was fed into the language model Vicuna (Llama 2-7b)[44] for language generation, performing VQA. For better multi-disease alignment, we leveraged the Retina Image Bank (2019-2023) to perform finetuning with fake QA's of 'Diagnosis:' and related condition keywords as answer, comprising 1,215 samples. Subsequently, we assessed the fine-tuned encoders on the open-set OphthalVQA dataset for zero-shot VQA testing. Alignment was conducted using the Low-Rank Adaptation (LoRA)[45] method with a batch size 8. This process spanned three epochs with an initial learning rate of 2e-5, and we incorporated cosine annealing for dynamic learning rate adjustments. Evaluation on OpthalVQA was carried out directly on the final epoch.

**Evaluation Metrics**

We employed the AUROC and AUPR metrics to assess the performance of classification tasks. These metrics gauge the classification effectiveness based on the receiver operating characteristics and precision-recall curves. When dealing with binary classification tasks like ocular disease diagnosis, we computed AUROC and AUPR in a binary context. For multi-class classification tasks such as five-stage DR and multi-class disease diagnosis, we calculated AUROC and AUPR individually for each disease class and then averaged them (macro) to derive the overall AUROC and AUPR scores.

Regarding VQA tasks, we utilized various classification-based metrics to evaluate performance, including the exact match score, F1 score, precision, recall, and the language-based metric known as BLEU[34].

For retrieval tasks, we used the metric Recall@K, which is the proportion of the data correctly retrieved among the top-K retrieved samples.

## Statistical Analysis

We calculated the mean performance and the standard error to derive the 95% confidence interval (CI) using 1.96 × standard error. We employed two-sided t-tests to compare EyeCLIP's AUROC with that of the most competitive models—CLIP, BioMedCLIP, PubMedCLIP, and RETFound—determining statistically significant differences.

## Data availability

We do not have permission to redistribute the datasets used for developing EyeCLIP. The downstream datasets can be accessed by referring to the original paper.

## Code availability

Code available at https://github.com/Michi-3000/EyeCLIP.

## Acknowledgments

We thank the American Society of Retina Specialists for providing the valuable Retina Image Bank.

## Author Contributions

D.S., J.Y., S.H., and M.H. conceived the study. D.S. and W.Z. built the deep learning model and ran experiments. M.H. provided data and computing facilities. D.S., W.Z., J.Y., S.H., and X.C. contributed to key data interpretation. D.S. W.Z, X.C. wrote the manuscript. All authors critically revised the manuscript.

## Competing interests

The authors declare no competing interest.

## Funding

Prof Mingguang He was supported by the Global STEM Professorship Scheme (P0046113). Dr Danli Shi was supported by the Start-up Fund for RAPs under the Strategic Hiring Scheme (P0048623) from HKSAR. The sponsor or funding organization had no role in the design or conduct of this research.

**Extended Figure 1**. Few-shot classification experiments. We investigated the label efficiency of different pretrained models in a few-shot setting, varying the number of training labels per class (nc = 1, 2, 4, 8, 16). For each nc, we sampled five different sets of training examples and trained a weakly supervised model. Boxes indicate quartile values, and whiskers extend to data points within 1.5× the interquartile range. EyeCLIP achieves significantly better performance (in terms of the mean AUPR of five runs) than other encoders for different sizes of training sets and across all datasets. AUPR = area under the precision-recall curve.

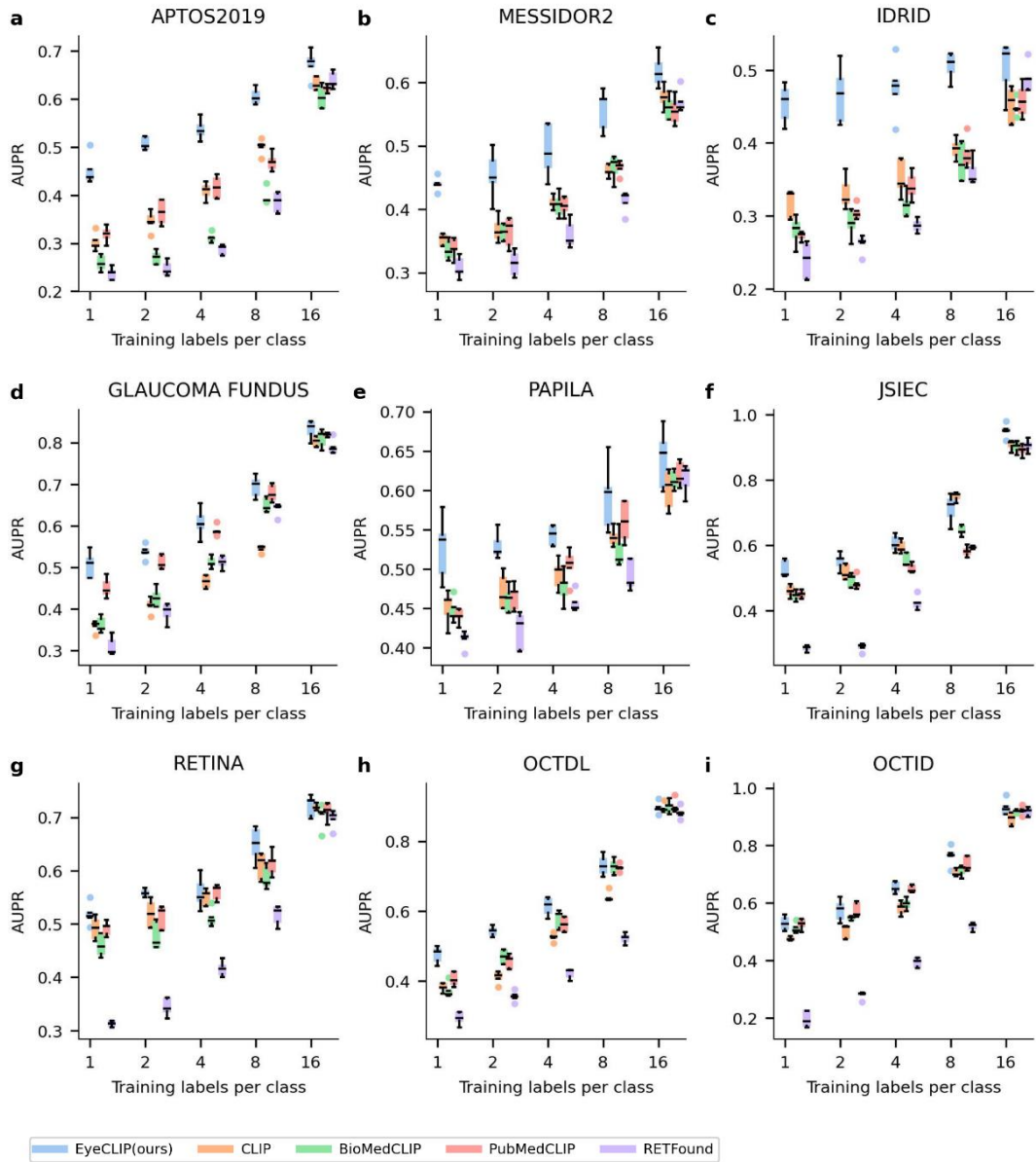

**Extended Figure 2**. More examples of cross-modal retrieval. CFP=color fundus photography, FFA=fundus fluorescein angiography.

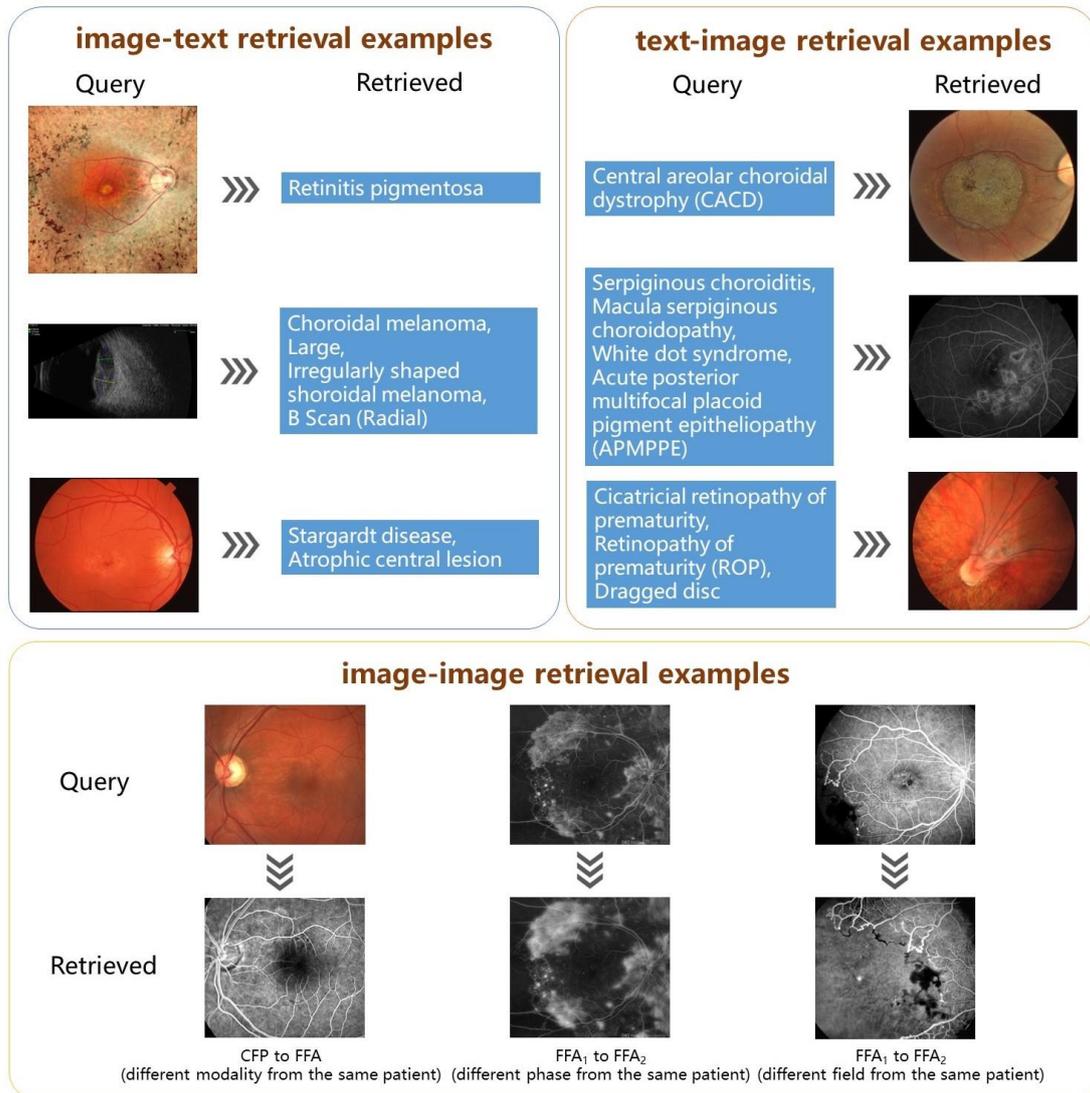